\documentclass{article}
\usepackage{ijcai20}

\usepackage{times}
\usepackage{soul}
\usepackage{url}
\usepackage[hidelinks]{hyperref}
\usepackage[utf8]{inputenc}
\usepackage[small]{caption}
\usepackage[dvipdfmx]{graphicx}
\usepackage{amsmath}
\usepackage{amsthm}
\usepackage{booktabs}
\usepackage{algorithm}
\usepackage{algorithmic}
\usepackage{bm}
\usepackage{multirow}
\usepackage{CJKutf8}

\newcommand{\citet}[1]{\citeauthor{#1} \shortcite{#1}}

\title{Bilingual Text Extraction as Reading Comprehension}

\author{
  Katsuki Chousa \And
  Masaaki Nagata \And
  Masaaki Nishino
  \affiliations
  NTT Communication Science Laboratories, NTT Corporation
  \emails
  \{katsuki.chousa.bg,masaaki.nagata.et,masaaki.nishino.uh\}@hco.ntt.co.jp
}

\begin{document}

\maketitle

\begin{abstract}
  In this paper, we propose a method to extract bilingual texts
  automatically from noisy parallel corpora by framing the problem as
  a token-level span prediction, such as SQuAD-style Reading
  Comprehension. To extract a span of the target document that is a
  translation of a given source sentence (span), we use either QANet
  or multilingual BERT. QANet can be trained for a specific parallel
  corpus from scratch, while multilingual BERT can utilize pre-trained
  multilingual representations.  For the span prediction method using
  QANet, we introduce a total optimization method using integer linear
  programming to achieve consistency in the predicted parallel spans.
  We conduct a parallel sentence extraction experiment using simulated
  noisy parallel corpora with two language pairs (En-Fr and En-Ja) and
  find that the proposed method using QANet achieves significantly
  better accuracy than a baseline method using two bi-directional RNN
  encoders, particularly for distant language pairs (En-Ja). We also
  conduct a sentence alignment experiment using En-Ja newspaper
  articles and find that the proposed method using multilingual BERT
  achieves significantly better accuracy than a baseline method using
  a bilingual dictionary and dynamic programming.
\end{abstract}

\section{Introduction}

Bilingual text extraction is the task of automatically extracting
parallel sentences of two languages from noisy parallel corpora. Both
the quantity and the quality of the bilingual texts used for training
are crucial for developing an accurate machine translation system.


\begin{CJK}{UTF8}{min}
\begin{figure}[ht]
    \centering
    \fbox{\begin{minipage}{\columnwidth}
        In meteorology, precipitation is any product of the
        condensation of atmospheric water vapour that falls under
        \textbf{gravity}. The main forms of precipitation include
        drizzle, rain, sleet, snow, graupel and hail... \\
        Q. What causes precipitation to fall? \\
        A. \textbf{gravity}
      \end{minipage}}
    \fbox{\begin{minipage}{\columnwidth}
        I said, ``Would you go to
        governments and lobby and use the system?'' He said, ``No, I'd
        take to the individuals.'' \textbf{It's all about the
          individuals.} It's all about you and me. It's all
        about partnerships... \\
        Q. 全ては個人についてのことであり \\
        A. \textbf{It's all about the individuals.}
      \end{minipage}}
    \caption{ Sample data of SQuAD dataset (upper) and the task of
      extraction from parallel corpora (lower).
    }
    \label{fig:crown}
\end{figure}
\end{CJK}

In this paper, we frame bilingual text extraction as a cross-language
span prediction problem similar to the SQuAD-style reading
comprehension task \cite{Rajpurkar_etal_EMNLP2016}. Figure
\ref{fig:crown} shows an example.  In SQuAD, given context C (a
paragraph from Wikipedia) and a question Q, the reading comprehension
system predicts an answer A as a span in the context.  Similarly, in
bilingual text extraction, given a target text as the context and a
source span as a question, the bilingual text extraction system
predicts a translation of the source text as the answer, which is a
span in the target text.

Recently, bilingual text extraction methods using neural networks have
gained popularity
\cite{Gregoire_Langlais_COLING2018,Artetxe_Schwenk_ACL2019,Yang_etal_IJCAI2019,Thompson_Koehn_EMNLP2019}.
These systems have two sentence encoders to obtain source and target
sentence embeddings and a scoring function to predict whether the two
sentences are parallel. Such approaches can be classified into two
categories: whether the two sentence embeddings are mapped into a
shared vector space \cite{Artetxe_Schwenk_ACL2019,Yang_etal_IJCAI2019}
or they are not \cite{Gregoire_Langlais_COLING2018}. The former type
uses cosine similarity with margin-based extensions for the scoring
function to solve the global inconsistency problem, while the latter
uses a feed-forward neural network for binary decisions (parallel or
not).  These approaches can also be classified by the type of encoder
used, such as a bi-directional Recurrent Neural Network
\cite{Gregoire_Langlais_COLING2018}, a Long-Short Term Memory network
\cite{Artetxe_Schwenk_ACL2019}, a Deep Averaging Network
\cite{Yang_etal_IJCAI2019}, or multilingual BERT
\cite{Yang_etal_IJCAI2019}.


Dual encoder approaches have also been used for reading comprehension
(question answering)
\cite{Seo_etal_ICLR2017,Yu_etal_ICLR2018,Devlin_etal_NAACL2019}.  As
for the encoder, BiDAF \cite{Seo_etal_ICLR2017} uses LSTM, while QANet
\cite{Yu_etal_ICLR2018} uses a combination of CNN and self-attention,
which is virtually equivalent to the Transformer
\cite{Vaswani_etal_NIPS2017}.  One of the architectural differences
between previous bilingual text extraction and reading comprehension
is that the latter adopts bidirectional cross attention
(context-to-query attention and query-to-context attention), which is
effective for capturing monolingual word-to-word interaction between
context and query.


We propose a novel bilingual sentence alignment method based on the
cross-language span prediction using reading comprehension
techniques. It essentially means that we use bidirectional cross
attention between context (target document) and query (source
sentence). We first used QANet \cite{Yu_etal_ICLR2018} because it is a
Transformer-based dual encoder with cross attention, which is more
powerful than a bidirectional RNN-based dual encoder
\cite{Gregoire_Langlais_COLING2018}.  We then used multilingual BERT
\cite{Devlin_etal_NAACL2019} because it is also a Transformer-based
encoder and its self-attention effectively includes cross attention
between source and target sentences when they are concatenated
as its input. Moreover, it can take full advantage of its powerful
pre-trained multilingual representations.

Since this span prediction method independently predicts target spans
for each source span, the target spans could have overlaps. Moreover,
because this method is asymmetric, the source-to-target predictions
could differ from the target-to-source predictions.  For the method
using QANet, we used an optimization method based on Integer Linear
Programming, which is a simplified version of a previous work
\cite{Nishino_etal_JNLP2016}.  For the method using multilingual BERT,
we simply averaged prediction probabilities from both directions.



We conducted two experiments to evaluate the proposed methods:
parallel sentence extraction from simulated noisy parallel corpora
(En-Fr and En-Ja) and sentence alignment for real newspaper articles
(En-Ja). We used a method using bi-directional RNN encoder
\cite{Gregoire_Langlais_COLING2018} as a baseline for parallel
sentence extraction because it can be trained for a specific parallel
corpus from scratch.  We found that the proposed method using QANet
achieves significantly better accuracy than the baseline, particularly
for distant language pairs (En-Ja).  We used a method using a
bilingual dictionary and dynamic programming
\cite{Utiyama_Isahara_ACL2003} as a baseline for sentence alignment
because it is commonly used for building publicly available
English-Japanese parallel corpora, including the shared task data for
NTCIR Patent Translation\footnote{http://ntcir.nii.ac.jp/PatentMT-2/}
and WAT (Workshop on Asian
Translation)\footnote{http://lotus.kuee.kyoto-u.ac.jp/WAT/}.  We found
that the proposed method using multilingual BERT achieves
significantly better accuracy than the baseline and the method using
QANet.

  
\section{Proposed Method}
\label{sec:proposed}

\subsection{Cross-Language Span Prediction by QANet}
\label{sec:extend_qanet}

The cross-language span prediction task is defined as follows: Suppose
we have a source document with N tokens
$F = \{ f_1, f_2, \dots, f_N \}$, and a target document with M tokens
$E = \{ e_1, e_2, \dots, e_M \}$.  Given a source text
$Q = \{ f_i, f_{i+1}, \dots, f_j \}$ that spans $(i, j)$ in the source
document $F$, the task is to extract target text
$R = \{ e_{k}, e_{k+1}, \dots, e_{l} \}$ that spans $(k, l)$ in the
target document $E$.

We first applied QANet \cite{Yu_etal_ICLR2018} to this task although
it is designed for reading comprehension, which is a monolingual span
prediction task.  To solve the span prediction task, QANet chooses a
span $(k, l)$ of target text $R$ corresponding to the source text $Q$
in the target document $E$ through the following five layers: Input
Embedding, Embedding Encoder, Context-Query Attention, Model Encoder
and Output Layers.

The embedding encoder layer is a stack of the block composed of
depthwise separable convolutions \cite{Kaiser_etal_ICLR2018},
self-attention with a multi-head attention mechanism
\cite{Vaswani_etal_NIPS2017}, and a feed-forward layer. It can be
considered equivalent to the Transformer encoder. The context-query
attention layer calculates context-to-query and query-to-context
attentions from $Q$ and $E$ to obtain weighted token vectors of $Q$
and $E$ by considering each other's information. The output layer
predicts the probability of each position $p_1$ and $p_2$ in the
target document becoming the start or end of an output span.  The
score of a span $\omega$ is defined as the product of its start and
end position probabilities.  The best span $(\hat{k}, \hat{l})$ is
chosen by maximizing the conditional probability, as follows:
\begin{align}
  \omega_{ijkl} &= p_1 (k | E, Q) \cdot p_2 (l | E, Q),
                  \label{eq:span_score} \\
  ( \hat{k}, \hat{l} ) &= \arg \max_{(k,l): 1 \leq k \leq l \leq M} \omega_{ijkl} .
                         \label{eq:span_argmax}
\end{align}
  
Furthermore, we need to determine whether the target text
corresponding to the source text exists since actual noisy parallel
corpora contain non-parallel sentences as noise.  For such a case, we
add an artificial token \textit{\textless NA\textgreater} at the
beginning of the target document, and if the model extracts only this
token, we assume that the corresponding target text does not exist.

We used a publicly available implementation of QANet
\cite{Yu_etal_ICLR2018} but made two important changes: First, we
applied Byte Pair Encoding
\cite{Sennrich_etal_ACL2016,Kudo_Richardson_EMNLP2018} after
tokenization to decrease out-of-vocabulary words. Source and target
vocabulary are shared, and the size of the shared vocabulary is set to
36,000. Secondly, we initialize the word (sub-word) embeddings with
uniform random values, while the original QANet used pre-trained Glove
word embeddings \cite{Pennington_etal_EMNLP2014} and converted all
unknown words into $<$UNK$>$ tokens.  In our preliminary
experiment, we found that the accuracy of the proposed model could be
improved by about 10\% using subword tokenization.

\subsection{Optimization of Predicted Spans by ILP}
\label{sec:optimization}

We define a score $\omega_{ijkl}$ for the target span $(k, l)$ given
source span $(i,j)$, which is obtained from the proposed model.  By
exchanging the source text and the target text in the model, we also
define a score $\omega'_{ijkl}$ for the same span pairs.  Since the
proposed model predicts a target span independently for the given
source span, there might be some overlap between predicted target
spans, even if the given source spans do not have overlap.  Moreover,
because the proposed model is asymmetric, the predictions from the
source text are very likely to be different from those from the target
text.  We need a total optimization method that can prevent spans from
overlapping and maximize the sum of predicted scores in both
unidirectional and bidirectional cases.

For total optimization of sentence alignment, dynamic programming
\cite{Gale_Church_CL1993} is commonly used although it assumes
monotonic alignment between the source and target sentences.  We use a
simplified version of a previous method \cite{Nishino_etal_JNLP2016}
because it can handle non-monotonic alignment and null alignment of
continuous segments using integer linear programming (ILP). We
formalize this problem as predicting a corresponding target span for a
given source span using a neural network and finding a maximally
non-overlapping pair of spans using ILP.



Let $d_{ijkl}$ be a pair of span $(i, j)$ in source text $F$
and span $(k, l)$ in target text $E$, and let $P$ be the set of all
possible pairs $d_{ijkl}$.  We can define a bilingual alignment $D$
for a document pair as a subset of span pairs $P$ ($D \subseteq P$),
where there is no overlap for any two span pairs in $D$. The ILP
formalization is as follows:
\begin{align}
  & \mathrm{Maximum}
  & \sum_{ijkl} \Omega_{ijkl} \ y_{ijkl} \label{eq:object_func} \\ 
  & \mathrm{Subject\ to}
  & y_{ijkl} \in \{0, 1\} \label{eq:cond_1} \\
  & \quad \sum_{i \leq x \leq j}{\sum_{kl}{y_{ijkl}}} \leq 1
  & \forall x: 1 \leq x \leq N \label{eq:cond_2} \\ 
  & \quad \sum_{ij}{\sum_{k \leq x \leq l}{y_{ijkl}}} \leq 1
  & \forall x: 1 \leq x \leq M \label{eq:cond_3} 
\end{align}
where $\Omega_{ijkl}$ is a score obtained from
$\omega_{ijkl}\text{ and }\omega'_{ijkl}$.  $y_{ijkl}$ is a variable
used to indicate whether the span pair $d_{ijkl}$ is included in the
alignment with $y_{ijkl} = 1$ showing that it is included.  Equation
(\ref{eq:cond_2}) guarantees that for each token in source text $F$,
there is at most one span pair $d_{ijkl}$ in the alignment that
includes the source token.  Equation (\ref{eq:cond_3}) guarantees the
same constraints for the target text $E$.  By combining the above two
constraints, each token in $E$ and $F$ is guaranteed to be included at
most once in the alignment.  We defined $\Omega_{ijkl}$ as follows:
\begin{equation}
    \Omega_{ijkl} = c \omega_{ijkl} + c' \omega'_{ijkl} .
\end{equation}
where $c \text{ and } c'$ are hyperparameters used to define the relative
importance of the source-to-target and target-to-source scores.  By
setting $c = 1, c' = 0$ or $c = 0, c' = 1$, the optimization becomes
unidirectional; by setting them to a positive value other than 0, it
becomes bidirectional.  In the experiment, we set $c$ to 1 and $c'$ to
the quotient of $max(\omega_{ijkl})$ divided by
$max(\omega'_{ijkl})$.

Since the references manually created for sentence alignment are based
on sentence boundaries, we searched for the nearest sentence
boundaries from the predicted span and regarded them as sentence-level
prediction. Furthermore, because the sentence-level units obtained in
this way may have more than one score, we filtered out the spans whose
score was less than $10^{-6}$ and took the average score of the
remaining spans for optimization.  We used ILOG CPLEX as a solver for
ILP.

\subsection{Cross-Language Span Prediction by BERT}

We then applied multilingual BERT \cite{Devlin_etal_NAACL2019} for the
cross-language span prediction. Although it is designed for such
monolingual language understanding tasks as question answering and
natural language inference, it works surprisingly well for the
cross-language span prediction task.

Since there are many null alignments in the sentence alignment of
comparable corpora, we adopted the SQuAD v2.0 format
\cite{Rajpurkar_elal_ACL2018}, which supports cases where there are no
answer spans to the question in the given context. We used the SQuAD
v2.0 model \cite{Devlin_etal_NAACL2019}, which adds two independent
output layers to pre-trained (multilingual) BERT to predict the start
and end positions in the context.  In the SQuAD model of BERT, first,
the question and the context are concatenated to generate a sequence
``[CLS] question [SEP] context [SEP]'' as input, where `[CLS]' and
`[SEP]' are classification token and separator token,
respectively. Then, the start and end positions are predicted as
indexes to the sequence. In the SQuAD v2.0 model, the start and end
positions are the indexes to the [CLS] token if there are no answers.
Since the original implementation of the BERT SQuAD model only outputs
an answer string, we modified it to output the answer's start and end
positions.

As for symmetrization (and optimization), we average the probabilities
of the best spans for each sentence in each direction. We treat a
sentence as aligned if it is completely included in the predicted
span. We then extract the alignments with the average probabilities
that exceed a threshold $\theta$. We set the threshold to 0.4 from the
results of preliminary experiments.  Although the span prediction of
each direction is made independently, we did not normalize the scores
before averaging because both directions are trained in a single
model.

As for null alignments, \citet{Devlin_etal_NAACL2019} used the
following threshold for the squad-2.0 model,
\begin{equation}
  \label{eq:non_null_threshold}
  \hat{s_{ij}} > s_{null} + \tau
\end{equation}
Here, if the difference between the score of the best non-null span
$\hat{s_{ij}}$ and that of a null (no-answer) span $s_{null}$ exceeds
threshold $\tau$, a non-null span is predicted.  The default value of
$\tau = 0.0$, and its optimal threshold is decided using the
development set. We used the default value because we assumed the
score of a null alignment is appropriately estimated since there are
many null alignments in the training data.

\section{Experiments on Noisy Parallel Corpora}
\label{sec:simulation_exp}

\subsection{Baseline method}
To show the effectiveness of the proposed approach, we first conducted
experiments on parallel sentence extraction. To evaluate the
performance of cross-language span projection using QANet without
total optimization, We extracted 1-to-1 bilingual texts from simulated
noisy parallel corpora and compared the results with those of an
earlier work \cite{Gregoire_Langlais_COLING2018} for a similar
language pair (En-Fr) and a distant language pair (En-Ja).

\citet{Gregoire_Langlais_COLING2018} first encode both source and
target sentences into two fixed-size continuous vectors using two
bidirectional RNNs (BiRNN).  From these sentence representations in a
shared vector space, they then estimated the conditional probability
that these sentences are parallel by applying a feed-forward neural
network. For the training dataset, they used parallel sentence pairs
in parallel corpora as positive examples. For negative examples, they
used negative sampling through sampling $m$ non-parallel sentences for
every positive sentence.

\subsection{Dataset}

\begin{table}[t]
  \centering
  \begin{tabular}{ccccc}
    \toprule
    {Corpus} & {Lang.} & \multicolumn{3}{c}{Number of Sentences} \\
             &         & Train & Valid. & Test \\
    \midrule
    Europarl & En-Fr & 500,000 & 1,000 & 1,000 \\
    KFTT & En-Ja & 440,288 & 1,166 & 1,160 \\
    IWSLT17 & En-Ja & 218,174 & 2,577 & 2,357 \\
    \bottomrule
  \end{tabular}
  \caption{Number of sentences for each corpus used in
    experiments.
  } 
  \label{table:corpora}
\end{table}

We used three parallel corpora composed of different language pairs:
Europarl
En-Fr\footnote{http://www.statmt.org/wmt15/translation-task.html},
KFTT\footnote{http://www.phontron.com/kftt/index.html}, and IWSLT17
En-Ja dataset\footnote{http://workshop2017.iwslt.org/}.  For the
Europarl dataset, we randomly chose 500,000 sentences for the training
set. Table \ref{table:corpora} shows their detailed statistics.

We created a dataset that has the same format as SQuAD v1.1 with
parallel corpus $P = \{(p_k^X, p_k^Y)\}_{k=1}^{K}$, where $X$ and $Y$
are arbitrary languages. To create the $k$-th data, we used a
source sentence $p_k^X$ as a query and a target sentence $p_k^Y$ as an
answer.  To generate context, we used negative sampling to insert $u$
negative sentences in front of and $(U-u)$ negative sentences behind
the output, where $U$ is the number of negative examples, and $u$ is a
random number from 0 to $U$.  On Europarl and KFTT, the negative
examples were sampled randomly.  On IWSLT17, we used sentences in
front of and behind the answer in the original document to keep the
context information.  By keeping this information, sentences that are
not parallel but similar to the query tend to appear in documents.  As
a result, the problem with context is more difficult than that without
context.

\subsection{Implementation Details}
We used a QANet model for SQuAD v1.1, which is implemented by
PyTorch\footnote{https://github.com/andy840314/QANet-pytorch-}. All
datasets were tokenized with
SentencePiece\footnote{https://github.com/google/sentencepiece}.  We
inserted nine negative sentences and filtered out from the training
set the queries and answers whose number of tokens was more than 100
tokens and the context was more than 1,000 tokens. The vocabularies
are shared between language pairs, and their total size is set to
36,000.

Adam \cite{Kingma_Ba_ICLR2015} was used for optimization, where the
minibatch size is 12.  We used a learning rate warm-up plan with the
inverse exponential increasing to $0.001$ during the first 1,000
steps.  The dropout probability was set to 0.1, the gradient clipping
was set to 5.0, and the coefficient value of weight decay was set to
$5\times 10^{-8}$.  Then, we chose the best parameter with the
smallest validation loss during 20 epochs.

As the baseline model, we used an implementation provided by its
authors\footnote{https://github.com/FrancisGregoire/parSentExtract}.
To generate negative examples, English and French sentences were
tokenized by the Moses
tokenizer\footnote{https://github.com/moses-smt/mosesdecoder/tree/master/scripts}
and the Japanese sentences were tokenized by
KyTea\footnote{http://www.phontron.com/kytea/}. Based on the original
paper, the number of negative examples was set to six, the noise ratio
was set to 0\%, and the threshold $\rho$ was set to 0.99.  In the test
set, the model first calculated the similarity between the sentences
of each document and the input sentence, and then extracted sentences
whose similarity was greater than or equal to a decision threshold
$\rho$.


For evaluation metrics, we used the token-level $F_1$ score and Exact
Match (EM) on the test sets.  The $F_1$ score was calculated against
the tokens of correct parallel sentence (span) pairs and predicted
parallel sentence (span) pairs. EM is defined as the accuracy of how
many predicted parallel sentences are exactly the same as the correct
parallel sentences.

\subsection{Results}

\begin{table*}[t]
  \centering
  \begin{tabular}{cccll}
    \toprule
    Corpus & Model & Direction & $F_1$ score & Exact Match \\
    \midrule
    Europarl
           & \cite{Gregoire_Langlais_COLING2018} & En-Fr & 94.56          & 94.50 \\
           &                                     & Fr-En & 94.60          & 94.60 \\
           & QANet                               & En-Fr & 98.57 (+4.01)  & 97.67 (+3.17) \\
           &                                     & Fr-En & 98.35 (+3.75)  & 98.17 (+3.57) \\
    \midrule
    KFTT
           & \cite{Gregoire_Langlais_COLING2018} & En-Ja & 81.84          & 81.38 \\
           &                                     & Ja-En & 81.52          & 80.78 \\
           & QANet                               & En-Ja & 98.06 (+16.22) & 96.25 (+14.87) \\
           &                                     & Ja-En & 97.43 (+15.91) & 92.23 (+11.45) \\
    \midrule
    IWSLT17
           & \cite{Gregoire_Langlais_COLING2018} & En-Ja & 62.73          & 62.70 \\
           &                                     & Ja-En & 62.35          & 62.31 \\
           & QANet                               & En-Ja & 95.58 (+32.85) & 86.50 (+23.80) \\
           &                                     & Ja-En & 96.99 (+34.64) & 93.44 (+31.13) \\
    \bottomrule
  \end{tabular}
  \caption{Experimental results with noisy parallel
    corpora. Direction indicates which language is a query and which
    is the answer. For example, ``En-Fr'' means that the query is
    written in English and the answer is written in French.}
  \label{tab:exp1_result}
\end{table*}

The experimental results in Table \ref{tab:exp1_result} shows that our
method using QANet is substantially better than the baseline for all
settings. It resulted in a higher $F_1$ score and EM than did the
baseline for both a similar language pair (En-Fr) with Europarl and a
distant language pair (En-Ja) with KFTT, even though it predicts a
span (sentence boundaries) by itself while the baseline uses given
sentence boundaries.  In a more difficult setting with IWSLT17, whose
documents contain contextual information, our method achieved
remarkable improvements of +34.64 points in $F_1$ score and +31.13
points in EM, compared with the baseline.

\section{Experiments on Comparable News Articles}
\label{sec:article_exp}


\subsection{Baseline method}
In the second experiment, we conducted a sentence alignment on actual
En-Ja newspaper articles.  Since newspaper articles contain
many-to-many alignments and null alignments (sentences with no
translations), the problem is substantially more complicated than the
one described in the previous subsection.


We used the method of a previous work \cite{Utiyama_Isahara_ACL2003}
as a baseline. To obtain article alignment, they first translated each
Japanese article into a set of English words using a bilingual
dictionary. They then used each English article as query and searched
for the most similar Japanese article in terms of BM25
\cite{Robertson_Walker_SIGIR1994}. They then aligned sentences in the
aligned articles using DP matching
\cite{Gale_Church_CL1993,Utsuro_etal_COLING1994} based on the
similarity measure SIM, which is defined as the relative frequency of
one-to-one correspondence between Japanese and English words obtained
from a bilingual dictionary. As a reliable measure for article
alignment, they used AVSIM, the average of SIMs obtained from the
sentence pairs in the article pair.  As a reliable measure for
sentence alignment, they used the product of article similarity AVSIM
and the sentence similarity SIM.


\subsection{Dataset}
We used a collection of newspaper articles and editorials from the
Yomiuri Shimbun and their translations published in The Japan News
(formerly the Daily Yomiuri), which is the newspaper's English
edition.  We purchased the newspaper's CD-ROMS for research
purpose\footnote{ https://database.yomiuri.co.jp/about/glossary}, and
created the manually and automatically aligned dataset as follows.

The manually aligned dataset consists of 157 bilingual document pairs
obtained by manually searching through 182 English documents for the
corresponding Japanese documents during two one-week periods
(2013/02/01-2013/02/07 and 2013/08/01-2013/08/07). It consists of 131
articles and 26 editorials. We manually aligned sentences for the 157
document pairs and obtained 2243 many-to-many alignments\footnote{We
  will make these annotations (both document alignment and sentence
  alignment) publicly available after our paper is published}.

Among the manually aligned data, we used the first 100 articles for
the training set, the next 15 articles for the test set, and the
remaining 16 articles as a future reserve. We also used the
automatically aligned data obtained using our implementation of the
previous method \cite{Utiyama_Isahara_ACL2003} as training data,
because the number of manually aligned documents and sentences is too
small.

For QANet, we used automatically aligned editorials as training data
because we found that the editorial pairs were highly accurate
sentence-by-sentence translations of each other. It is probably
because they represented the official opinions of the newspaper
company.  From 19,113 Japanese editorials and 11,434 English
editorials from 1989 to 2016, we automatically extracted 11,414
bilingual documents and obtained 299,178 many-to-many alignments.  We
used all automatically aligned editorials (except those used for the
development set) and the first 100 articles and all 26 editorials in
the manually aligned dataset for the training set, and we used the
final 50 editorials in the automatically aligned data for the
development set.

For multilingual BERT, we used all articles and editorials in 2012, It
consists of 24,293 Japanese documents and 4,878 English documents. We
automatically obtained 663 editorial pairs and 2,989 article pairs,
and then extracted 16,409 and 40,373 many-to-many alignments,
respectively. Articles contain a fair amount of non-parallel sentences
because some English articles are abstracts of Japanese articles and
sometimes additional explanations are added to the English articles
for readers who are not familiar with Japan and Japanese culture.
Since the SQuAD v2.0 model of multilingual BERT explicitly models null
alignments, we assumed it is better to use articles for training the
model.

For the QANet model, we treat the manually aligned data and
automatically aligned data equally.  We used an entire document as
context and removed alignments in the context having non-continuous
spans.  We made negative examples to learn null alignments as follows:
For editorials, we sampled random sentences that are not included in
the context as negative examples. For articles, we sampled sentences
with no alignment relations as negative examples. Negative examples
selected for editorials amounted to as much as 10\% of the total
sentences.

For the SQuAD v2.0 model of multilingual BERT, we first used the
automatically aligned data for fine-tuning of 5 epochs. We then used
the manually aligned data for fine-tuning of another 5 epochs. We also
removed alignments with non-continuous spans from the training data.

\subsection{Implementation Details}

We used BERT-Base, Multilingual Cased (104 languages, 12-layer,
768-hidden, 12-heads, 110M parameters, November 23, 2018) in our
experiments\footnote{https://github.com/google-research/bert}. We used
the script for SQuAD v2.0 as is. The parameters are as follows: train
batch size is 6, learning rate is 3e-5, number of training epochs is
5, maximum sequence length is 384, maximum query length is 158,
maximum answer length is 158, and doc stride is 64.  Since the number
of input tokens for BERT is limited to 512, it is difficult to
accommodate both query (source sentence) and context (target document)
in the window size. To avoid out-of-memory errors, we have to lower
the maximum sequence length and batch size further.

The SQuAD v2.0 model of BERT has adopted a sliding windows approach,
where a window for the context, whose length is maximum sequence
length minus maximum query length minus two, slides with a stride of
doc stride. Since an answer can appear in multiple windows, the score
with "maximum context," which is defined as the minimum of its left
and right context, is taken (the sum of left and right context will
always be the same).  A sentence longer than either maximum query
length or maximum answer length is simply truncated.



For the baseline method, we used our implementation of an earlier work
\cite{Utiyama_Isahara_ACL2003}.  Sentences were split using sentence
boundary symbols \footnote{\begin{CJK}{UTF8}{min}！, ？,
    and。\end{CJK} for Japanese and !, ?, :, ;, and . for English}
with additional rules, and these were tokenized by MeCab-UniDic for
Japanese and TreeTagger for English.  We used the EDR
Japanese-to-English dictionary, EDR English-to-Japanese dictionary,
and EDR Technical Term Dictionary for the experiment.  The number of
entries is 483,317 for Japanese-to-English and 367,347 for English to
Japanese.

The evaluation was done based on the number of sentence pairs
extracted by the alignment methods. We used Precision/Recall/$F_1$
score as the evaluation measure for sentence alignment.

\subsection{Results}
\begin{table}[t]
  \centering
  \begin{tabular}{crrr}
    \toprule
    Model & Precision & Recall & $F_1$ \\
    \midrule
    \cite{Utiyama_Isahara_ACL2003} & 54.1 & 50.0 & 51.9 \\
    \midrule
    QANet (Ja-En)                  & 56.3 & \textbf{67.3} & 61.3 \\
    QANet (En-Ja)                  & 57.2 & \textbf{67.3} & 61.8 \\
    QANet (Ja-En) + ILP            & \textbf{72.5} & 65.8 & \textbf{69.0} \\
    QANet (En-Ja) + ILP            & 64.8 & 59.6 & 62.1 \\
    QANet (Bidi) + ILP             & 67.3 & \textbf{67.3} & 67.3 \\
    \midrule
    BERT (Ja-En) & 83.5 & 70.6 & 76.5 \\
    BERT (En-Ja) & 86.0 & 69.9 & 77.1 \\
    BERT (Bidi) & \textbf{86.4} & \textbf{74.6} & \textbf{80.1} \\
    \bottomrule
  \end{tabular}
  \caption{Experimental results using actual newspaper articles.
    Bold indicates the highest value for QANet and BERT.}
  \label{tab:exp2_result}
\end{table}
  
Table \ref{tab:exp2_result} shows the results.  Our method using QANet
and bidirectional ILP optimization is significantly better (15.4 $F_1$
points) than the baseline. Our method using multilingual BERT and
symmetrization is significantly better (12.8 $F_1$ points) than that
using QANet.

Japanese-to-English and English-to-Japanese predictions have about the
same accuracies for both QANet and multilingual BERT. ILP optimization
improves precision at the cost of recall. For QANet, we think
bidirectional ILP optimization is better in terms of the balance
between precision and recall although the $F_1$ of Ja-En
uni-directional ILP optimization is higher.  In multilingual BERT, a
simple combination (symmetrization) of two directional predictions
improves both precision and recall, which results in 3 $F_1$ points
improvement.

\section{Related Works}

Previous methods for sentence alignment are based on
context-independent similarity of source and target sentences such as
sentence length \cite{Gale_Church_CL1993}, bilingual dictionaries
\cite{Utsuro_etal_COLING1994,Utiyama_Isahara_ACL2003,Varga_etal_RANLP2005},
and sentence embeddings
\cite{Gregoire_Langlais_COLING2018,Artetxe_Schwenk_ACL2019,Yang_etal_IJCAI2019,Thompson_Koehn_EMNLP2019}.
They usually use dynamic programming, which assumes that the
alignments are monotonic. On the contrary, the proposed method
considers the context of a target sentence and can handle
non-monotonic alignments.


\citet{Gregoire_Langlais_COLING2018} proposed a method to extract
parallel sentences by using a dual encoder based on bi-directional
RNN, and they achieved high accuracy in sentence alignment between
English and French, but their experiment was done only on synthesized
data.  \citet{Artetxe_Schwenk_ACL2019} and \citet{Yang_etal_IJCAI2019}
proposed parallel corpus mining methods based on multilingual sentence
embedding in a shared vector space. Both works used pre-trained
encoders and a scoring function using cosine distance with some
margin-based extension. Moreover, both works reported state-of-the-art
results in the BUCC shared task on parallel corpus mining
\cite{Zweigenbaum_etal_BUCC2018}.

Since the targets of previous works on parallel corpus mining using
neural networks were mainly among European languages, it is not clear
whether these methods work effectively on distant language pairs such
as English and Japanese.  Furthermore, these methods were tested in an
easier setting than that of the real problem.  For example, the BUCC
shared task \cite{Zweigenbaum_etal_BUCC2018} assumes sparse 1-to-1
sentence alignment in synthesized bilingual documents and the WMT
corpus filtering task \cite{Koehn_etal_WMT2018} assumes that the
sentences are already aligned. We applied our method to sentence
alignment of real newspaper articles in a distant language pair and
showed its effectiveness.

Recently, \citet{Thompson_Koehn_EMNLP2019} proposed a sentence
alignment method, called Vecalign, which uses bilingual sentence
embeddings \cite{Artetxe_Schwenk_ACL2019} and recursive DP
approximation. They used a German-French test set and achieved
state-of-the-art results. Comparing our method with theirs remains
future work. It should be noted that we can use their outputs for
fine-tuning our model before using the manually created data to
fine-tune it further.


\section{Conclusion}
In this paper, we proposed a novel sentence alignment method based on
cross-language span prediction, which can be implemented either by
QANet or multilingual BERT.  Future works include investigating the
best practice for combining manually and automatically aligned data
because the amount of manually aligned data for training is usually
limited.






\bibliographystyle{named}
\bibliography{ijcai2020}

\end{document}